\documentclass[11pt,letterpaper]{article}

\usepackage[utf8]{inputenc} 
\usepackage[T1]{fontenc}    
\usepackage[letterpaper, margin=1in]{geometry} 
\usepackage{amsmath, amssymb, amsfonts}       
\usepackage{graphicx}       
\usepackage{microtype}      
\usepackage{hyperref}       
\usepackage{abstract}       
\usepackage{graphicx}
\usepackage{float}
\usepackage{caption}
\usepackage{enumitem}
\usepackage{subcaption}
\usepackage{amsmath}
\usepackage{amssymb}
\usepackage{adjustbox}
\usepackage{tabularx}
\usepackage{array}
\usepackage{balance}
\usepackage{makecell}
\usepackage{xcolor}
\usepackage{algorithm} 
\usepackage{pgfplots}
\pgfplotsset{compat=newest}
\usepackage{array}
\usepackage{algorithmic}
\usepackage{soul}
\usepackage{pifont}
\usepackage{multirow}

\newcommand\blfootnote[1]{%
  \begingroup
  \renewcommand\thefootnote{}\footnote{#1}%
  \addtocounter{footnote}{-1}%
  \endgroup
}
\definecolor{mygreen}{HTML}{009900}%
\definecolor{myred}{HTML}{CC0000}%
\definecolor{myblue}{HTML}{0000CC}%
\definecolor{mysky}{HTML}{0075FB}%
\definecolor{myyellow}{HTML}{F4D03F}%
\definecolor{myyellow1}{HTML}{FFA500}%
\definecolor{myblack}{HTML}{212F3D}%
\definecolor{myorange}{HTML}{ff4500}%
\definecolor{mybrown}{HTML}{B22222}%
\definecolor{myblack}{HTML}{000000}
\definecolor{mygreen1}{HTML}{056608}%
\definecolor{mypink}{HTML}{FF1493}
\definecolor{mypurple}{HTML}{663399}%

\hypersetup{
    colorlinks=true,
    linkcolor=blue,
    filecolor=magenta,      
    urlcolor=cyan,
    citecolor=blue,
}

\title{\textbf{Resource-Efficient Multimodal Fusion for Early Rice Blast Forecasting in Sustainable Agriculture}}
\author{Shubhajyoti Das$^1$, Pritam Bikram$^1$, Arindam Biswas$^1$, Samrat Paul$^2$, \\Parimal Sinha$^2$, and
        Bhargab B. Bhattacharya$^1$\\ \\
        $^1$ Indian Institute of Engineering Science and Technology, Shibpur\\
        $^2$ ICAR-Indian Agricultural Research Institute, New Delhi\\ \\
        Email(s): dasshubhajyoti@gmail.com, pritambikram6@gmail.com,
        barindam@gmail.com,\\
        samrat16.sp@gmail.com,
        sinhapath@gmail.com, 
        bhargab@isical.ac.in}

\date{24$^{th}$ November 2026}

\begin{document}

\maketitle
\begin{abstract}
Resource-efficient artificial intelligence has become increasingly important for sustainable agriculture, where timely disease forecasting must be achieved under computational and energy constraints. Early estimation of rice blast can enhance food security by optimizing crop management practices while reducing excessive pesticide use. However, existing frameworks often suffer from high computational overhead and limited capability to model the complex spectral–temporal dependencies present in heterogeneous meteorological and remote sensing data. To address these challenges, this paper proposes a lightweight multimodal fusion framework comprising an attentive positional encoding-based Transformer and a convolutional neural network (CNN)-embedded gated recurrent unit (GRU) for efficient spectral–temporal representation learning. The attention-enhanced positional encoding assigns adaptive importance to sequential meteorological tokens, enabling effective modelling of temporal characteristics. In parallel, the lightweight CNN extracts discriminative spectral features, while the GRU captures temporal dynamics from remote sensing imagery acquired over two time-instances. A multimodal fusion layer integrates these heterogeneous representations to forecast rice blast risk $5$\textendash$10$ days before symptom manifestation across diverse agro-climatic zones. Extensive experiments demonstrate that the proposed framework outperforms state-of-the-art methods, achieving a validation accuracy of $96.93\%$ and an $F1$-score of $95.79\%$. In addition to its predictive performance, the lightweight multimodal architecture substantially reduces computational complexity, inference latency, and carbon emissions, making it an energy-efficient solution for large-scale disease forecasting. The proposed framework provides a practical foundation for resource-efficient artificial intelligence in sustainable agriculture, supporting timely disease management and environmentally responsible crop production.
\end{abstract}
\blfootnote{The work done by B. B. Bhattacharya was supported by the ANRF Prime Minister Professorship Grant at IIEST, Shibpur.}
\section{Introduction}\label{sec 1}
Crop disease monitoring and analysis is a necessary part of sustainable farming, aimed at controlling fungicides, pesticides, labour, and equipment \cite{wadhwa2024generalizable}. Excessive usage of agrochemicals harms the environment and undermines natural farming. However, crop losses are exacerbated by uncontrolled protection practices, leading to extensive chemical use \cite{contreras2026enhancing}. Crop protection techniques, including early symptom detection, are essential for efficient disease control \cite{ou2023application}. Advanced technologies that support early disease monitoring and forecasting processes are in high demand in smart agricultural industries aiming to maximize productivity while maintaining a sustainable environment \cite{ma2024early}. Sensor and imagery data could be ideal for sustainable farming practices and early disease management \cite{yang2020remote}. Hence, identifying favourable periods before symptom onset is useful for determining the exact timing of intervention to prevent the spread of disease and boost the country's health and economy. Rice farming is unsettled by leaf blast disease (\textit{Pyricularia oryzae}), which causes severe damage to the leaves, leaf blades or sheaths, nodes, neck, and even panicles as the plant reaches physiological maturity in worldwide \cite{zhao2022study, kaur2024development}. It is difficult to identify early disease in the field since infected plants do not exhibit overt signs on their higher stems and leaves in the early stages of infection \cite{xie2024hyperspectral}. 
Meteorological data, such as temperature, humidity, rainfall, and hourly sunshine, have a significant impact on blast disease estimation \cite{nettleton2019predicting, malicdem2015rice}. Temperature is a critical element that directly or indirectly controls numerous aspects of disease development \cite{bregaglio2010integrated,viswanath2017simulation}.
However, only pneumatic factors are insufficient to recognize disease patterns.\\
Satellite-driven remote sensing data can measure crop growth across multiple spectral bands (optical, thermal, and microwave), and temporal vegetation index analysis may help monitor crop phenology \cite{guan2017shared, liu2022detecting,dutta2023satellite, yu2021early}. Plant diseases can be detected early using hyperspectral or multi-spectral imaging, as well as RGB (red, green, blue) imaging \cite{almoujahed2022detection, das2023deep, tian2023disease}.
Therefore, both temperature (meteorological data) and spectral index are potential variables in forecasting disease stages in the field.
\graphicspath{{Figure/}}
\begin{figure*}[]
    \centering
    \includegraphics[width=0.85\textwidth]{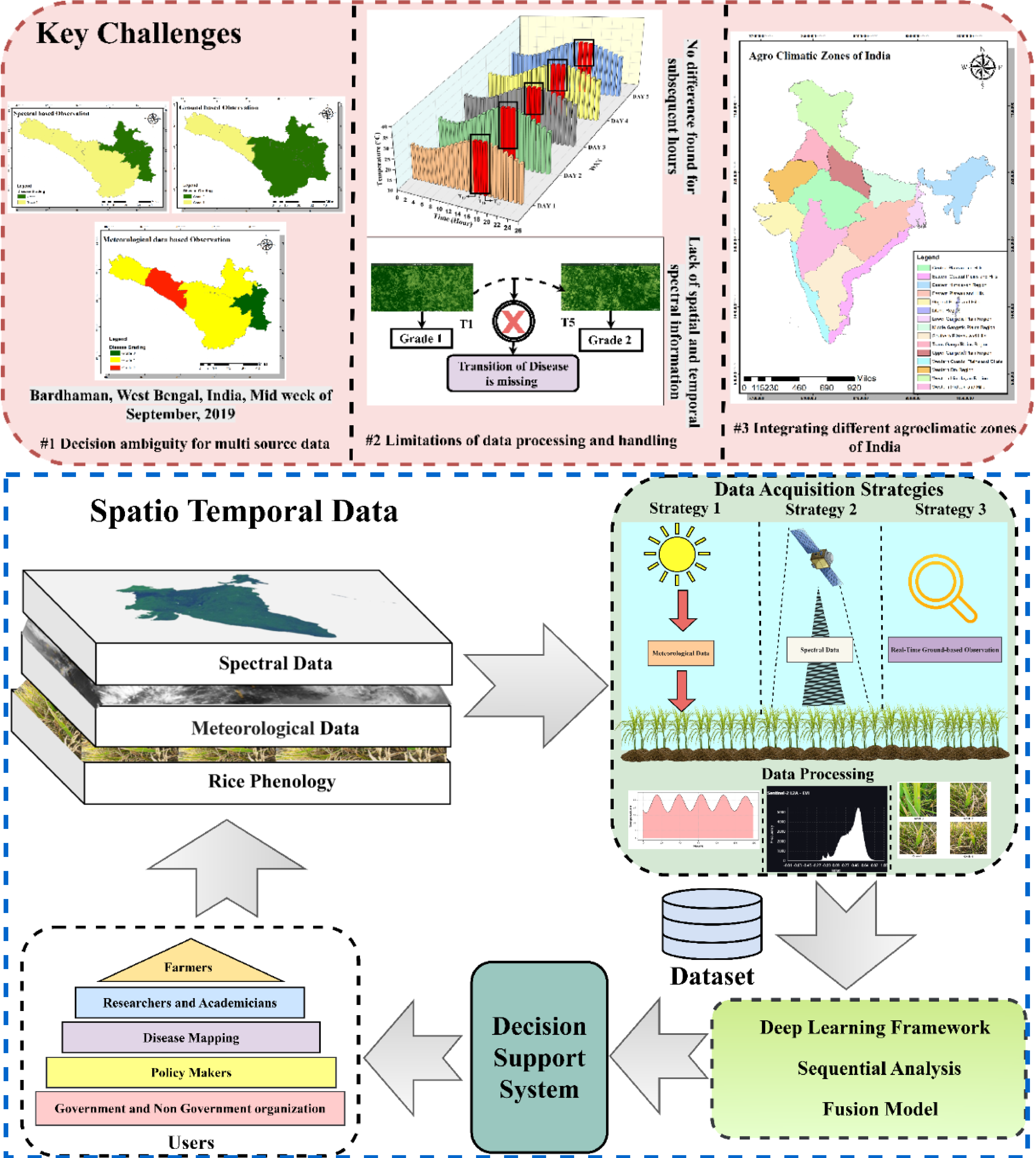}
    \caption{Key challenges of early disease prediction model: $(1)$ multi-source data shows decision ambiguity due to diverse inputs, $(2)$ spatiotemporal meteorological and spectral data poses a lack of efficient data representation, $(3)$ heterogeneous agroclimatic conditions.Proposed Schematic workflow of multi-modal data collection and transmission, considering all limitations.}
    \label{fig 101}
\end{figure*}
CNNs and their variants can extract more significant features from an image than previous techniques for colour and texture extraction \cite{gong2024integrating}. 
Remote sensing-based spectral images exhibit a nonlinear relationship between pixels, providing vibrant information about the environmental features of an area. CNNs have great potential for extracting long-range spatial relationships from spectral images \cite{gao2024bridging}. 
An optimized CNN-based model with layer stacking enabled feature extraction from spectral images in a structured and coherent manner \cite{das2025multilayer}.
Recurrent Neural Networks (RNNs) are very adept at extracting features from time series data. Long Short-Term Memory (LSTM) and GRU are RNN versions that address the issues of gradient vanishing and gradient explosion inherent in standard RNNs \cite{wang2023deep}. Sun \textit{et al.} suggested that CNN-based LSTM could perform spatiotemporal feature extraction for optimized monitoring of crop phenology \cite{sun2020multilevel}. GRU presented a simplified structure that retains LSTM's capacity to store long-term states. GRU networks have fewer parameters, shorter training times, and lower computational complexity. Due to these advantages, the CNN-enabled GRU network performed efficiently in prediction tasks utilizing spatiotemporal features \cite{chen2022short}. Jiang \textit{et al.} used meteorological and remote-sensing data to develop a phenology-based LSTM model for estimating maize production in the United States \cite{jiang2020deep}. Transformers and their variants are among the many cutting-edge algorithms that have emerged to handle time-series forecasting tasks \cite{jiang2024applicability}. Lightweight transformer networks can improve predictive performance in high-dimensional meteorological data analysis \cite{11352790}. To analyze long sequential data, the informer model was employed for sequential climatic aspects \cite{liu2022rice}. A multi-scale spatiotemporal transformer network is well-suited for capturing the nonlinear relationships in meteorological data \cite{li2024multi}. The spatial Transformer, equipped with a positional encoding module, leverages local and global geographic similarities in weather systems to capture multi-scale spatial accessibility in meteorological graphs \cite{ji2024spatio}. For such diverse data analysis, huge computational parameters need to be adjusted to enhance the model's performance. Large computational overhead causes a carbon footprint and increased electricity consumption, thereby jeopardizing sustainability.\\
Nonetheless, as the extent of the several locations continues to expand and the importance of managing more complex disease situations escalates, numerous basic issues remain in the existing early disease prediction in sustainable farming, as described in Figure \ref{fig 101}. The major research gaps are listed below:
\begin{enumerate}[label=\roman*.]
\item Previous research relies on spectral imaging technology and meteorological data analysis separately for rice blast disease forecasting. Although both datasets pose challenges for disease analysis due to nonlinear relationships among various ecological conditions, shortcomings in previous research suggest analyzing them in ways that allow each to contribute to its own implications.
\item Hourly data with a few differences; data redundancy occurs during data processing, as described in the challenges part of Figure \ref{fig 101}. Only the spectral spatial data failed to capture the disease transition phase. Meanwhile, the practical implementation of advanced deep learning models has limitations due to their complex architectures and the large number of learnable parameters, both of which increase their carbon footprint.
\item Several agroclimatic zones represent complex, nonlinear data for disease modelling. Therefore, a zone-based model harms the model's generalizability and robustness, thereby affecting crop protection techniques.
\end{enumerate}

In view of this, the intelligent agricultural systems require environmental sustainability in addition to agronomic results \cite{11498480}. This study proposes a lightweight learning framework for early recognition of the different stages of rice blast disease by combining sequential meteorological and spectral features, aiming to be eco-friendly. Sequential temperature data and spectral variables, including the Enhanced Vegetation Index (EVI), are used to estimate blast infection stages with two deep learning models: the positional attentive transformer and the CNN-GRU.
The attentive positional encoded lightweight transformer is proposed to extract long-range sequential features, improving the significance of temporal meteorological information. The CNN-GRU model efficiently extracts spectral-temporal information over the same consecutive time frame. This work aims to establish a robust, generalizable framework by leveraging a fusion technology that integrates an attentive positional-encoding transformer and a CNN-GRU network for prognostication of rice blast disease using climate factors, particularly temperature, and remotely sensed spectral variables associated with EVI.\\
The main contributions are as follows:
\begin{enumerate}
    \item The attention mechanism in the transformer network's positional encoding emphasizes each token, maintaining a coherent representation of hourly temperature. It enhances the robustness of the transformer model for extracting contextual relationships between tokens. The lightweight transformer encoder is implemented to extract long-range dependencies of the meteorological data. A multi-level convolution network embedded with GRU is introduced to capture spectral-temporal characteristics from spectral images. 
    \item The hierarchical fusion of heterogeneous data helps identify disease occurrence and its relationship with the environment. The attentive hourly temperature escalates disease progression, increasing the importance of crucial hours with their attention factor. The spectral-temporal information provided insights into the previous and current states of environmental conditions with respect to disease occurrence. The fusion learning framework collects and processes disease states with significant characteristics.       
    \item Evaluate the stability and generalization of the proposed lightweight fusion technology embedded with the transformer and CNN-GRU model over many years, and explore spatiotemporal variations of the disease in different regions of India while maintaining sustainable farming.    
\end{enumerate}
\section{Methodology}\label{sec 2}
\subsection{Study Area and Data Acquisition}
\subsubsection{Study Area}
For leaf blast observation, $22$ locations were selected across the $15$ agroclimatic zones available in India. Blast incidence during September to November $(2018-2023)$ was collected from selected locations. The temperature in the above-mentioned areas ranges from 40\textdegree C to 11\textdegree C during the experimental period. A plant pathologist recorded the samples with visible blast symptoms on the leaves. The disease was observed by selecting $4 - 12$ random fields at each location. The experts classify disease occurrences by lesion size, density, and number \cite{bock2021phytopathometry}, dividing them into three categories: low, moderate, and severe disease intensity.  
\subsubsection{Meteorological Data}
Weather-related variables were gathered to comprehend the relationship between meteorological data and rice blast disease \cite{garrett2022climate, lee2023deep}. Temperature is a key factor among other weather factors due to its interdependence with the blast's infectious stages \cite{viswanath2017simulation}. The data were collected from the NASA archives for the specific experimental regions and time, including the maximum temperature (MaxT) and minimum temperature (MinT).  
The synthetic generation of the hourly temperature \cite{bregaglio2010integrated} for the disease latent period can be expressed as:
\begin{multline}
\label{eq 2}
T_{(\text{air}-h)}=\frac{(MaxT + MinT)}{2} + \frac{(MaxT - MinT)}{2} \times \cos{(0.2618 \times (h - \text{TimeVar}))}
\end{multline}
where TimeVar is the hour of the day corresponding to MaxT and h is the hour of the day. The generated data is structured into five-day intervals for the sequential predictive model.
\begin{figure}[!h]
    \centering
    \includegraphics[width=0.9\textwidth]{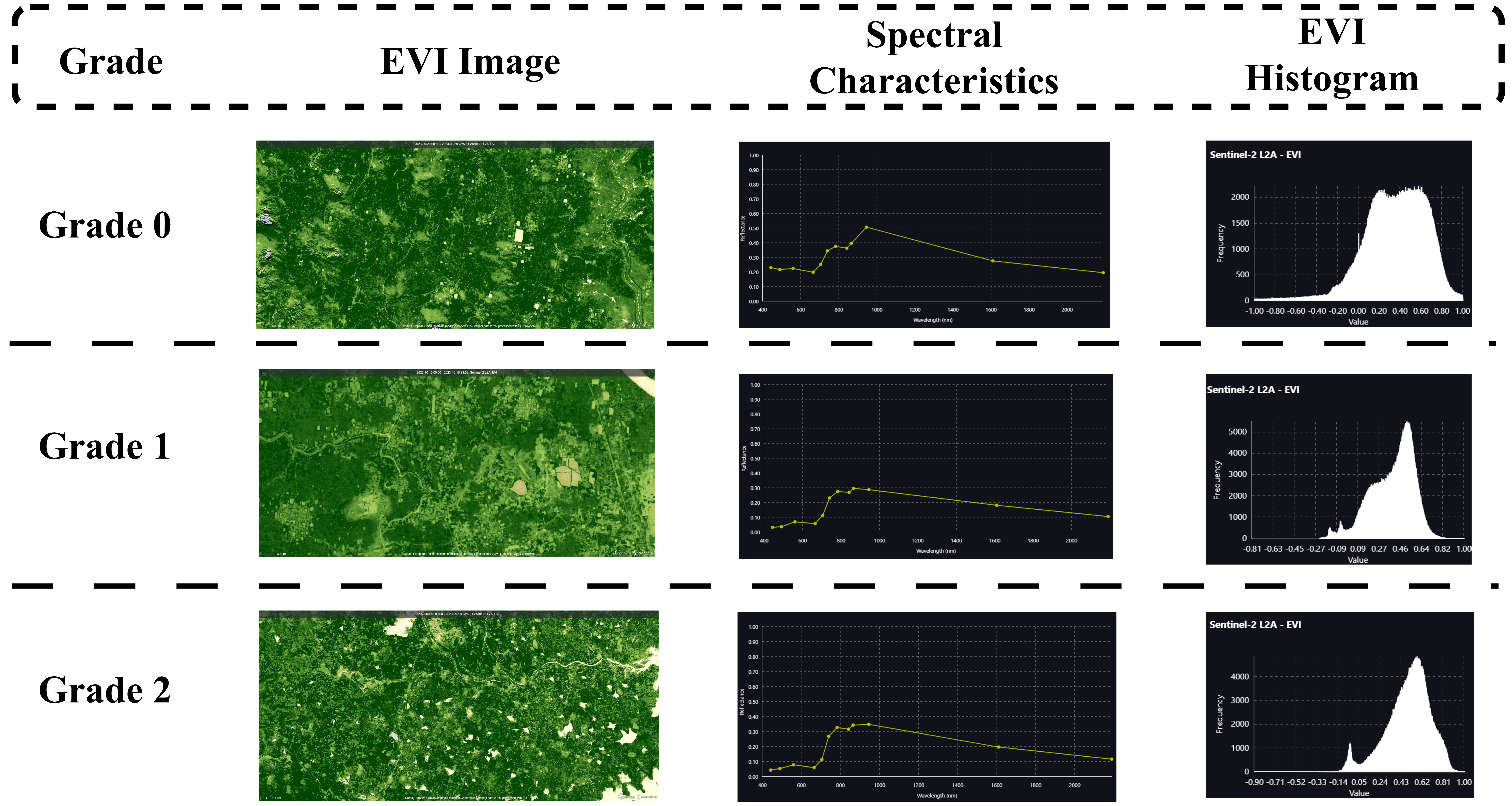}
    \caption{Grade-wise spectral distribution of rice blast disease. Spectral characteristics represent the reflectance value of bands $B1$ ($443$ nm), $B2$ ($490$ nm), $B3$ ($560$ nm), $B4$ ($665$ nm), $B5$ ($705$ nm), $B6$ ($740$ nm), $B7$ ($783$ nm), $B8$ ($842$ nm), $B8A$ ($865$ nm), $B9$ ($945$ nm), $B11$ ($1610$ nm), $B12$ ($2190$ nm). The histogram of EVI indicates the interval of low to high vegetation density as determined by the computed EVI values for every pixel.}
    \label{fig 4}
\end{figure}
\subsubsection{Remotely sensed Spectral data}
The remote sensing data utilized in this study were obtained from the Sentinel $2$ satellite. 
Sentinel $2$ product EVI spectral index, comprised of near-infrared (NIR), red edge (R), and blue edge (B), is a highlighting factor for rice blast disease recognition from space. The spectral range of EVI ($2.5 \times [(NIR-R)/(((NIR + 6.0) \times (R -7.5) \times B)+1)]$) is generally in the range of $600$ nm – $900$ nm \cite{tian2023disease, das2025multilayer}.
Figure \ref{fig 4} shows the EVI spectral characterization and band histogram of various grades of rice blast. The developed dataset is graded based on ground observations. 
\subsection{Proposed Framework}
\begin{figure*}[]
    \centering
    \includegraphics[width=\textwidth]{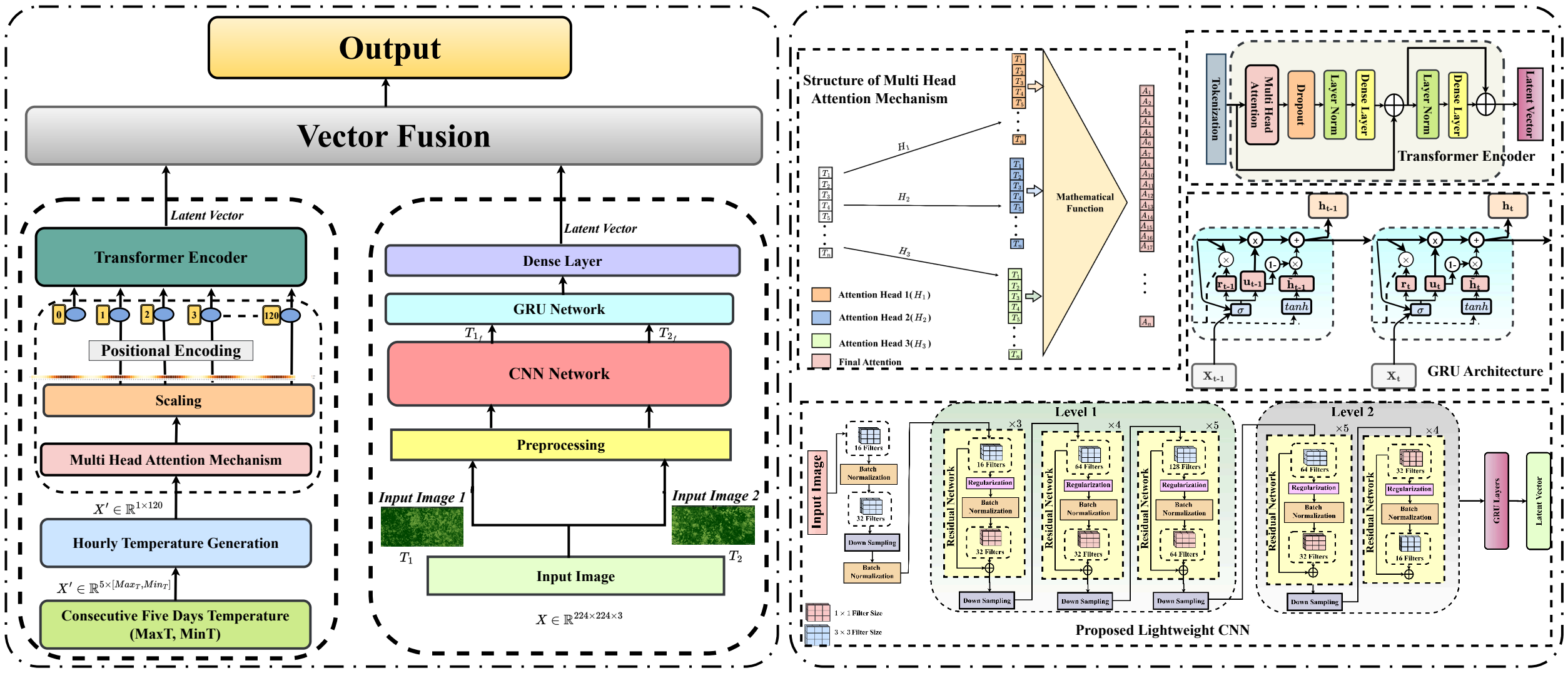}
    \caption{Description of the proposed framework and its components. CNN and GRU networks extract spatio-temporal spectral information from two consecutive days (Total $ 5$ days). Transformer network captures coherent features from meteorological data.}
    \label{fig 5}
\end{figure*}

In this research, a hierarchical fusion framework is proposed that combines an attentive, lightweight positional transformer with a CNN-embedded GRU network to analyze sequential temperature and spectral data. A systematic approach is employed to construct, optimize, and evaluate the proposed model's efficacy for forecasting blast disease. The pipeline of the proposed model is demonstrated in Figure \ref{fig 5}.
\graphicspath{{Figure/}}
\subsubsection{Transformer network for meteorological data}
The Transformer module is constructed using the lightweight encoder mechanism and attentive positional tokenization to capture the contextual information of time-series data. Consequently, the length of the time-series data is $120$. The dimension of one input sequence is ($1 \times 120$). 

The multi-head attention mechanism (Figure \ref{fig 5}) is designed to apply more weight to all tokens. The primary objective is to apply an attention mechanism on tokens to efficiently encode sequential data. Weight functions are generated in the attention mechanism. Tokens are multiplied with weight functions to initiate the weighted features, such as $X^H = W_a \times X + b$, where $W$ and $b$ are the weight and bias. $X$ denotes the total number of tokens. To generalize all weighted characteristics, they are concatenated and multiplied by another weight function, then passed through a softmax activation. The mathematical expression of the function is:
\begin{align}
    A_i=Softmax[((Concat(M_x) \times W_t  )+b_t ) ]
\end{align}
All heads are merged to form a multi-head attention mechanism. Where $M_x$ denotes one head attention features and $W_t$ and $b_t$ describe weights and bias simultaneously. 
The generated $A_i$ is an attention feature that is combined with the input tokens to achieve the actual attentive features for positional encoding. Attentive feature is derived as:
\begin{align}
    X_A=Softmax\left[\sum_{i=1}^{n} A_i \times X_{input}\right]
\end{align}
$X_A$ is a crucial factor in the transformer's positional encoding of the token. The factor helps reorganize the token and provides generalization based on disease occurrence. It is expressed mathematically as:
\begin{align}
    PE(pos,2i) = \sin{(pos/10000^{2i/d_{model}})\times X_A}
\end{align}
\begin{align}
    PE(pos,2i+1) = \cos{(pos/10000^{2i/d_{model}}) \times X_A}
\end{align}
There are two components to the positional encoding: $PE (pos, 2i)$ and $PE (pos, 2i+1)$ for even-indexed and odd-indexed dimensions, respectively. Here, $pos$ indicates the location of a token in the sequence, and $i$ is the index of the positional encoding dimension, which begins at $0$. In the Transformer model, $d_{model}$ is the feature dimension. Two different dimensions are depicted by the mathematical function $f(X_{\bar{A}})=[PE (pos, 2i), PE(pos,2i+1)]$. Following positional encoding, the position encoding vectors are appended to the token vectors as input to the Transformer Layer. 
The proposed lightweight transformer encoder comprises a multi-head self-attention module, Layer Normalization ($LN$), Dropout ($\delta$) and Dense Layer. The self-attention component generates attention scores for each temporal characteristic, allowing the model to focus on meaningful context while suppressing noise. Multi-head self-attention utilizes sets of learned attention weights, enabling the model to capture a wide range of connections and dependencies in the data \cite{wang2024lightweight}. The input vectors are embedded into the self-attention component, which consists of query $(Q)$, key $(K)$, and value $(V)$ to compute the attention weight matrices$\left(Softmax\left[\left(\frac{Q \cdot K^T}{\sqrt{C}}\right)\times \delta \right] \times V\right)$ for a particular dimension $D$.
Each token learns its relationship to other tokens in the sequence via self-attention. This enables the model to create complex representations by combining outputs from multiple heads and extracting information from multiple perspectives. The residual connection is introduced in the module that stabilizes the learning weights and prevents overfitting. It also helps to process the main information with the processed one. The structural description of the proposed encoder (Figure \ref{fig 5}) can be mathematically described as:
\begin{flalign}
    \hat{X} &= SelfAttention(f(X_{\bar{A}})) \nonumber \\
    X^{\prime(l)} &= LN(\delta \cdot \hat{X}) \nonumber \\
    X^{\prime(l)} &= \sum w \times X^{\prime(l)} + b \nonumber \\
    X_{con} &= X^{\prime(l)} \oplus f(X_{\bar{A}}) \nonumber\\
    \hat{X}^{\prime(l)} &= \sum w \times LN( X_{con}) + b \nonumber \\
    y &= X_{con} \oplus \hat{X}^{\prime(l)}
\end{flalign}
The transmitted vector is passing through two encoding layers with two heads. The processed information from the self-attention mechanism progresses across the dropout layer with a dropout rate of $50$\%. Layer normalization is then applied to stabilize the output distribution. The first dense layer with $120$ neurons and the “GELU” activation function remolds the input into a specific dimension. The reshaped information is concatenated with the input to preserve the gradient descent situation. Then, layer normalization is applied before fetching into the second dense layer with the same configuration as the first dense layer. Consequently, the information is merged with the first residual connection information to serve the latent features. The enhanced model delivers more accurate and robust features for further analysis. 
\begin{table*}[!ht]
\setlength{\tabcolsep}{10pt}
\renewcommand{\arraystretch}{1.5}
\centering
\captionsetup{justification=centering}
\caption{Evaluation of models using training accuracy, validation accuracy, precision, recall, F$1$ score and AUC values. The complexity of different models is also evaluated using trainable parameters and FLOPs.}
\resizebox{\textwidth}{!}{
\begin{tabular}{|c|c|c|c|c|c|c|c|c|}
\hline
 \textbf{ \makecell{Model\\ variation}} & \textbf{\makecell{Training\\ Accuracy}} & \textbf{\makecell{Validation\\ Accuracy}} & \textbf{Precision} & \textbf{Recall} & \textbf{\makecell{F$1$\\ score}} & \textbf{\makecell{AUC\\ values}} & \textbf{\makecell{Trainable\\Parameters\\(M)}} & \textbf{\makecell{FLOPs \\(G)}}\\ \hline
 Transformer\cite{ji2024spatio} & $85.32$ & $75.40$ & $76.65$ & $75.56$ & $76.20$ & $75.80$ & $8.2$ & $18.6$ \\ \hline
 CNN-GRU\cite{liu2026improved} & $82.32$ & $76.40$ & $77.65$ & $77.26$ & $78.20$ & $77.45$ & $1.8$ & $1.5$\\ \hline
Lstm + CNN-Lstm\cite{santoshi2026lddhybridnet} & $85.60$ & $78.32$ & $79.20$ & $78.60$ & $78.89$ & $79.80$ & $5.2$ & $3.24$ \\ \hline
GRU\cite{wang2023deep} + CNN-Lstm\cite{santoshi2026lddhybridnet} & $88.12$ & $79.25$ & $80.26$ & $80.80$ & $80.52$ & $82.50$ & $5.04$ & $3.18$ \\ \hline
GRU + CNN-GRU\cite{liu2026improved} & $90.5$ & $82.8$ & $84.86$ & $83.2$ & $84.02$ & $85.6$ & $2$ & $1.80$ \\ \hline
SpectralFormer\cite{wu2025advanced}& $90.5$ & $88.5$ & $87.52$ & $88.15$ & $87.90$ & $88.45$ & $4.9$ & $3.10$ \\ \hline
 \textbf{\makecell{Transformer + \\CNN-GRU\\
(Proposed Model)
}}& \textbf{96.50} & \textbf{96.93} & \textbf{95.60} & \textbf{96.00} & \textbf{95.79} & \textbf{97.32} & \textbf{2.15} & \textbf{3.01}\\ \hline
\end{tabular}}
\label{tab 4}
\end{table*}
\subsubsection{CNN-GRU framework for spectral images}
A novel CNN-enabled GRU network is designed to analyze consecutive spectral images for spectral-temporal data. CNN is implemented for an efficient feature extraction strategy from the complex nonlinear spectral features of the image. However, it is difficult to extract spatial spectral features due to the complexity of pixel distributions. Hence, 
the proposed CNN network is configured with convolution layers, batch normalization (BN), strides, $L2$ regularization, and GELU activation functions. Convolution layers with various filters and kernel sizes are arranged in a two-level structure that enables efficient feature extraction (Figure \ref{fig 5}). BN is applied after each convolution layer to normalize and optimize the processed features \cite{11397421}. Strides were used to downsample the large input in a specific manner. To address overfitting, a regularization strategy is essential. It requires putting a penalty term to be reduced into the loss function equal to the sum of squared parameters or weights of a model \cite{das2025multilayer}. The GELU activation function \cite{raza2024multiclass} uses a percentile-based weighting system. For this reason, the GELU activation function is used in each convolution layer. The mathematical expression of the proposed network structure is derived as follows:
\begin{align}    
    X_{l} &= GELU\left[\left (\sum_{i \in Mj}^{n} X_i \times K_{ij}\right) + b_j \right] \times L2 \nonumber \\
    X_F &= BN[X_l]
\end{align}
Where $X_i$ represents the input features, $K_{ij}$ is the kernel, and $b_j$ is the bias. $X_l$ denotes the output of convolution features, and $X_F$ is the final output feature from the convolution layer, and $L2$ describes the regularization factor. \\
However, scaling the trained weights alone is not enough to account for the complex, non-linear relationships in the pixel distribution. 
The primary purpose of the residual structure in the model is to preserve the original features and enhance model efficacy, thereby reducing vanishing gradient issues \cite{li2025residual}. The first layer combination demonstrates $X_l \in I^{(K\times C)}$, and the following layer combination depicts $X_{(l+2)}\in I_{l+2}^{K \times C}$ ($K$ and $C$ as kernel and channel simultaneously). The residual network implies $X_l \oplus X_{(l+2)}$, which maintains the capability of the optimized network. In this study, kernel sizes of $3\times3$ and $1\times1$ are used. The first level of the network uses $16$, $32$, $64$, and $128$ filters, and the second level uses $64$, $32$, and $16$ filters.\\ 
Another component of this framework is the GRU network, used for temporal analysis of spectral images. The primary attribute of the GRU (Figure \ref{fig 5}) is its gating mechanism, which comprises the update gate and reset gate. 
GRU fully leverages the inherent properties of time-series data and can capture dependencies across multiple time scales \cite{wang2023deep} using both gates.
The mathematical description of the GRU structure can be described as:
\begin{align}
    U_t &= \sigma (W_U \cdot[h_{t-1},X_t]) \nonumber \\
    r_t &= \sigma (W_r \cdot[h_{t-1},x_t]) \nonumber \\
    \tilde{h_t} &= tanh(W_{\tilde{h}} \cdot [r_t \otimes \mid h_{t-1}, X_t]) \nonumber \\
    h_t &= (1-U_t) \otimes h_{t-1} + U_t \otimes \tilde{h_t}
\end{align}
$U_t$ and $r_t$ are the update and reset gates, respectively. The sigmoid function, represented by $\sigma$, has an output that ranges from $0$ to $1$, $X_t$ is the input, $\tilde{h}_{t-1}$ is the output state of the previous time, $h_t$ is the output value at the current time, $\tilde{h_t}$ represents the hidden state information of the current time, $\otimes$ is matrix elements multiplication, and $W_r$, $W_U$ and $W_{\tilde{h}}$ are the weights of the reset gate, update gate, and current memory content, respectively. Three GRU units are used for this analysis, with each set to $128$ units. One dropout layer is used to regularize the weights with a dropout rate of $0.2$.\\
The time series ${{x_1,x_2,x_3, \dots x_n}}$ is used as input image for the interval of $5$ days. Therefore, the actual input image passes through the CNN layers, is transformed into $(6\times6\times16)$- dimensional features, and then passes through the GRU network to generate spatial-temporal relationships from images of the disease condition. 
\subsubsection{Fusion of Heterogeneous Data}
The processed meteorological features ($X_{M}$) and spectral-temporal features ($X_{S}$) are concatenated in the fusion layer. The fusion layer consists of multiple fully connected layers with $128$ and $64$ neurons, respectively. The fused features ($X_M \oplus X_S$) are processed by the fully connected layers. The features provide class-wise probabilities using the “Softmax” function, and the maximum value yields an output for each observation. The fusion archictecture is defined as:
\begin{flalign}
    X_{fusion} &= X_M \oplus X_S\nonumber\\
    X_{D1} &=  \sum^{128}_{i=1} w_i \times X^{fusion} + b\nonumber\\
    X_{D2} &= \sum^{64}_{i=1} w \times X_{D1} + b\nonumber\\
    Output &= Softmax(X_{D2})
\end{flalign}
\section{Experimental Results and Discussion}\label{sec 4}
\subsection{Experimental Design}
In this research, all model training, validation, and testing are performed on a device configured with a $12^{th}$ Gen Intel(R) Core(TM) $i9-12900$ $2.40$ GHz processor and $64$ GB RAM. The device also consists of an NVIDIA RTX $A2000$ GPU with $12$ GB for the experiments. The dataset is divided into $7:2:1$ for training, validation, and testing. The learning rate was set to $0.0001$ for $100$ epochs, with a batch size of $8$ and “Adam” optimizer. Training and validation accuracies, precision, recall, and F$1$-score are evaluated to assess model performance. Additionally, computational efficiency is discussed to support the sustainability of the proposed model while maintaining a green environment. 
\subsection{Model Interpretation}
\begin{figure}[!h]
    \centering
    \includegraphics[width=0.75\textwidth]{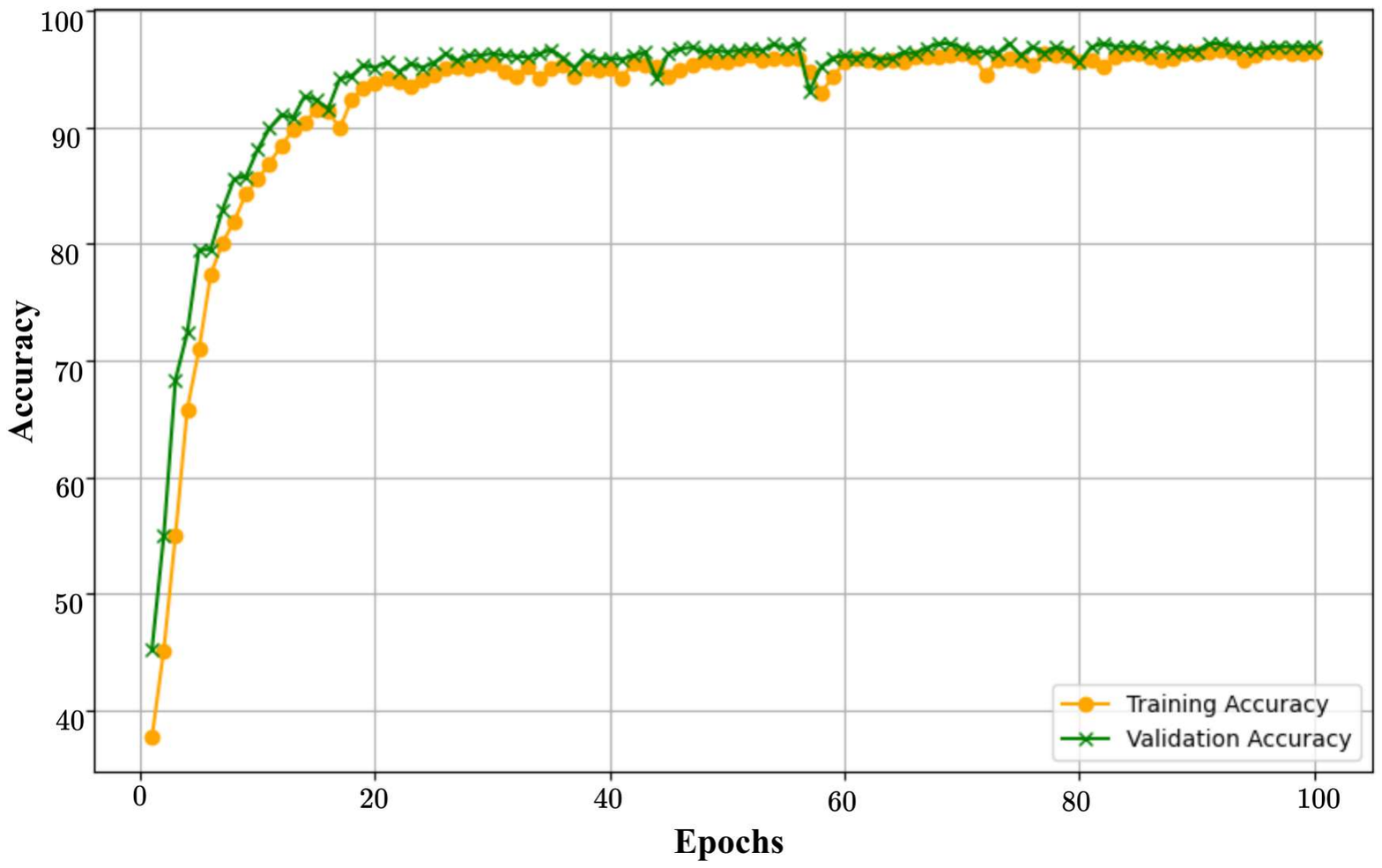}
    \caption{Graphical representation of training accuracy and validation accuracy of the proposed model.}
    \label{fig 9}
\end{figure}

The classification results of the proposed model are compared with those of other models, as shown in Table ~\ref{tab 4}.
All models show significant increases in accuracy curves and decreases in loss curves as epochs increase. Differences between training and validation accuracies are observed during model training. Significant differences between those metrics are observed to mitigate the overfitting and underfitting issues \cite{das2023deep}. The training process of the proposed model is depicted in Figure \ref{fig 9}, which shows minimal differences between training and validation accuracies. The graph also shows that the difference between the two metrics decreases across consecutive epochs.
Table \ref{tab 4} shows that different models provide different classification results across various metrics. The proposed model achieved $96.5\%$, $96.93\%$, $95.6\%$, $96.00\%$, $95.79\%$, and $97.32\%$ for training accuracy, validation accuracy, precision, recall, F$1$ score, and AUC values, respectively. The model showed significant performance differences of $3.1\%$, $8.33\%$, $9.60\%$, $9.68\%$, and $9.67\%$ in training accuracy, validation accuracy, precision, recall, F$1$ score, and AUC values, respectively. The CNN with GRU can extract critical features from spectral-temporal relationships, while the transformer model is intended to effectively incorporate significant meteorological features. Furthermore, the statistical analysis of the proposed model is evaluated for different disease-level estimates. The classification report of the model is depicted in Table \ref{tab 5}. It shows different grades ranging from $93.00\%$ to $95.00\%$, $95.00\%$ to $97.00\%$, and $94.00\%$ to $96.00\%$ for precision, recall, and F$1$ scores separately.
\begin{table}[!htbp]
\caption{Classification report of the proposed model.}
\centering
\begin{tabular}{|c|c|c|c|}
\hline
 \textbf{Grade} & \textbf{Precision}  & \textbf{Recall}  & \textbf{F1 Score}  \\ \hline
Grade $0$ & $94.32$  & $96.20$ & $95.75$  \\ \hline
Grade $1$ & $93.30$  & $95.21$  & $94.54$  \\ \hline
Grade $2$ & $95.10$  & $97.50$ & $96.28$  \\ \hline
\end{tabular}
\captionsetup{justification=centering}
\label{tab 5}
\end{table}

\subsection{Model Evaluation }
The classification results have shown that the fusion approach achieves significant performance gains in disease analysis.
The fusion data are tested considering the rice growing season of August to October from $2018$ to $2023$. 
The experiments have shown that a high misclassification rate occurs in August and September, whereas it is lower in October, using only meteorological data. The misclassification rates for August and September are $26.26\%$ and $26.84\%$, respectively. The rate is relatively low for October, with $19.42\%$. The analysis of spectral data are also evaluated  
\begin{figure}[htbp] 
     \centering
     \begin{subfigure}[b]{0.31\linewidth} 
         \centering
         \includegraphics[width=\linewidth]{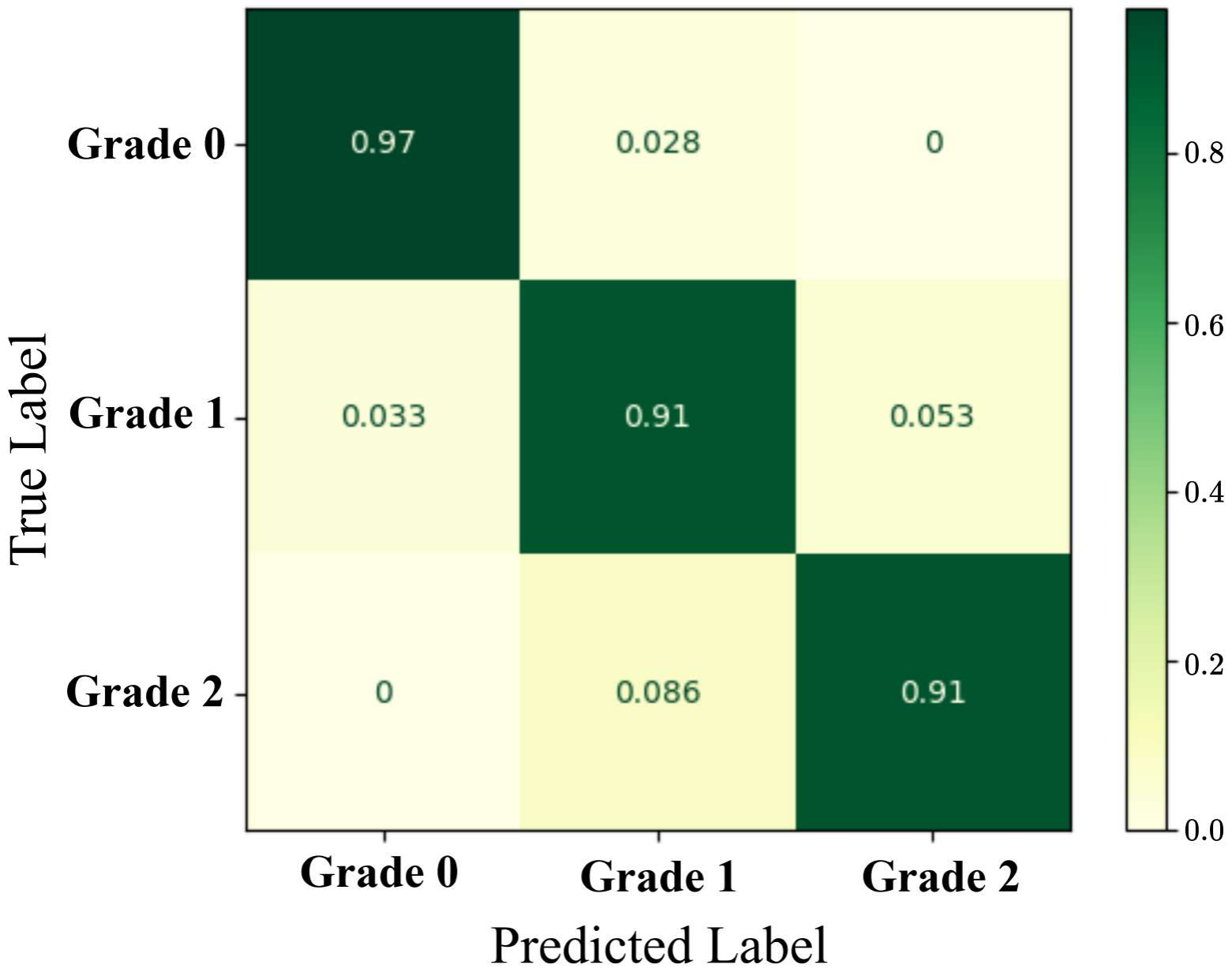}
         \caption{Confusion matrix of  the month of August}
         \label{12a}
     \end{subfigure}
     \hfill
     \begin{subfigure}[b]{0.31\linewidth}
         \centering
         \includegraphics[width=\linewidth]{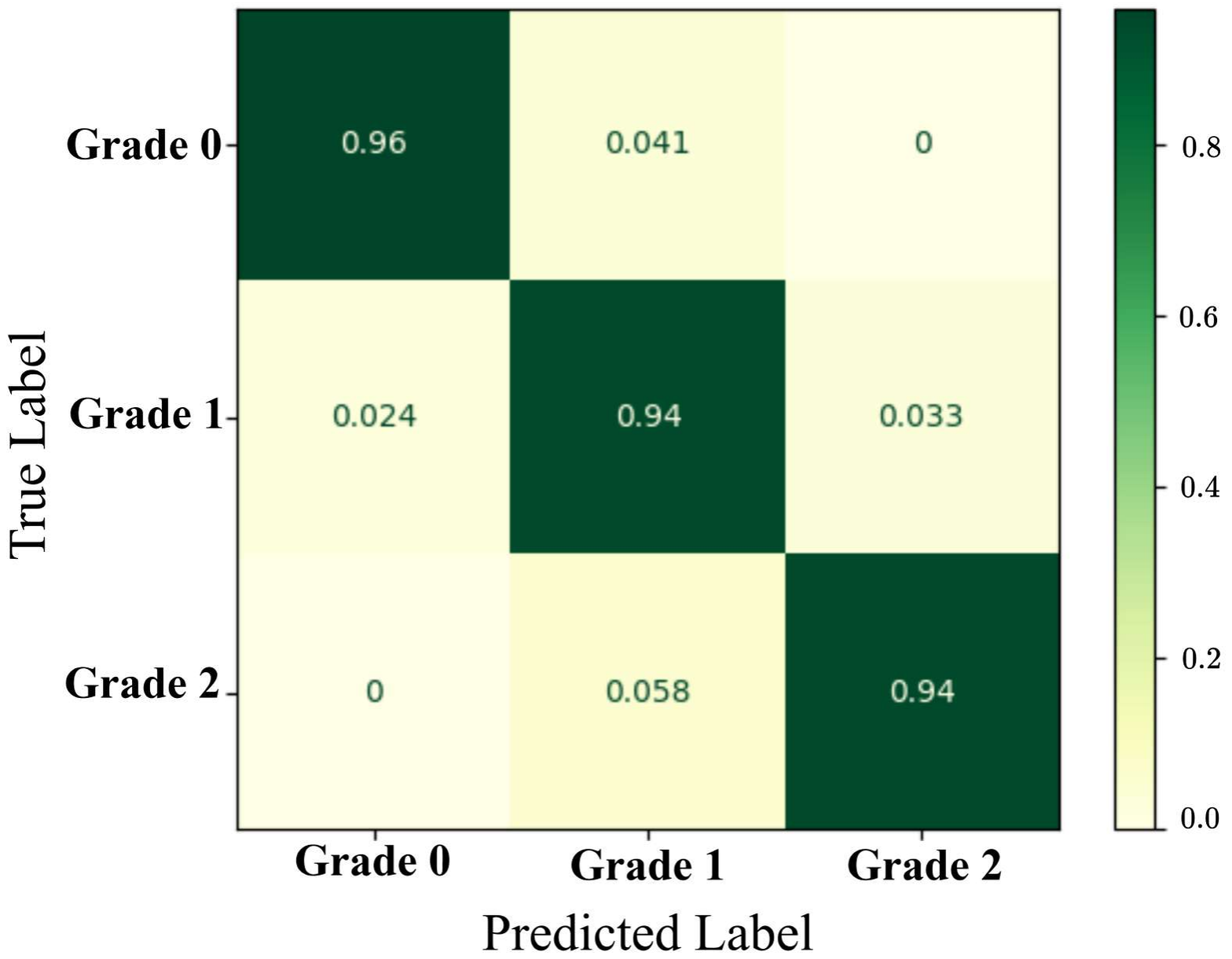}
         \caption{Confusion matrix of  the month of September}
         \label{12b}
     \end{subfigure}
     \hfill
     \begin{subfigure}[b]{0.31\linewidth}
         \centering
         \includegraphics[width=\linewidth]{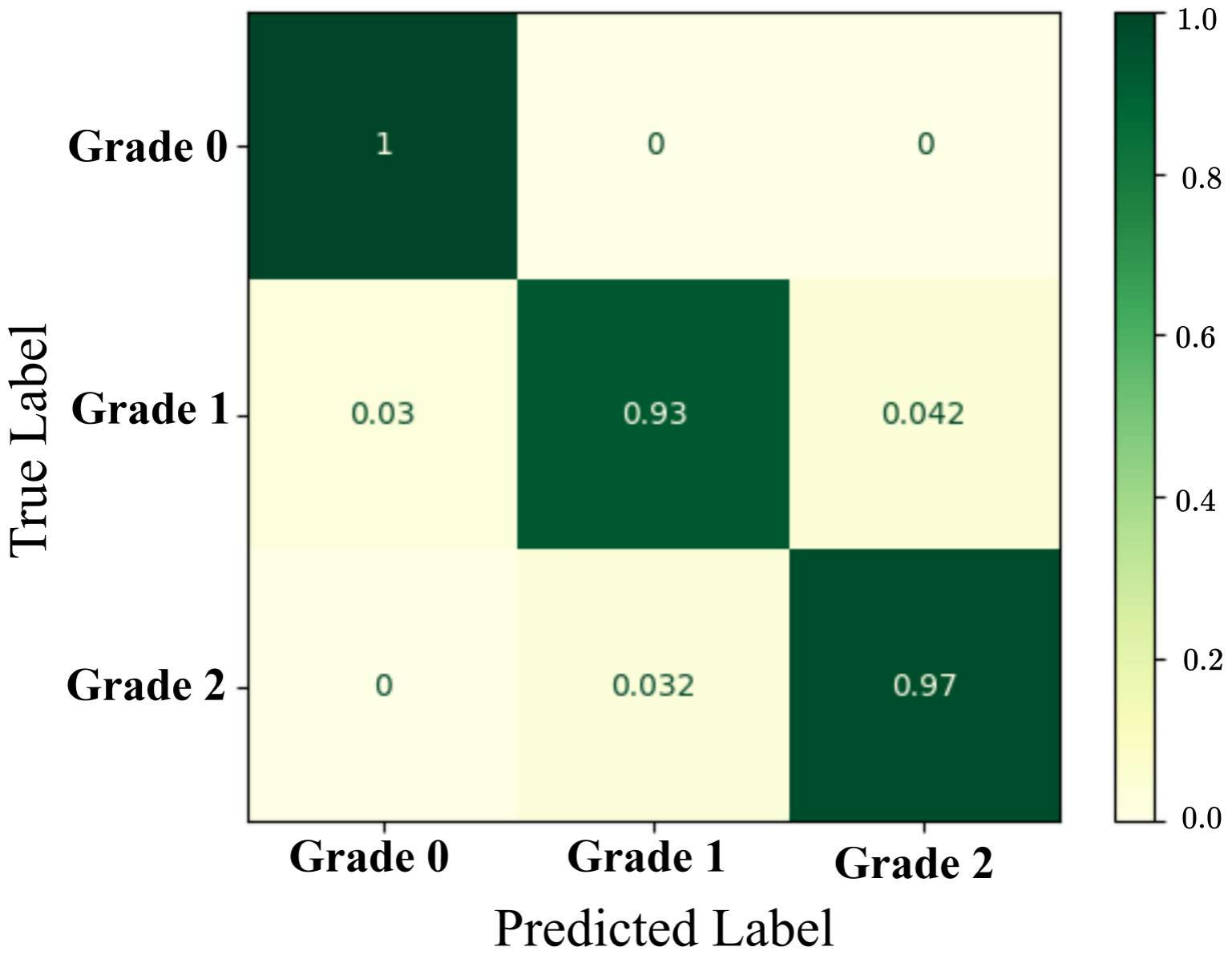}
         \caption{Confusion matrix of the month of October}
         \label{12c}
     \end{subfigure}
        \caption{Fusion model of confusion matrices for different months.}
        \label{fig 12}
\end{figure}
and observes high misclassifications in August and September. The rate is $29.28\%$ and $27.89\%$ for August and September, respectively. The rate is low for October at $15.85\%$. In contrast, both independent methods are unsuitable for early disease estimation. Since the fusion approach is expected to improve the performance of the prediction model. Figure \ref{fig 12} describes the confusion matrices of the fusion model for August, September, and October. Therefore, utilizing both modules in a combinational technology is best for all situations. Nevertheless, it is important to observe the distribution of features across different disease grades. Figure \ref{fig 13} demonstrates the distribution of features of various grades for different months.
Differences in features were apparent in the PCA scatter plot, where each class was distinctly separated from the others. The PCA scatter plot shows that some features are present in the common regions of different classes. However, maximum features are distinguished with large distances. Thus, the fusion approach is more efficient and effective for analyzing rice blast disease.\\
The results showed that classification accuracy significantly increased to $93.36\%$, $94.80\%$, and $95.70\%$ for August, September, and October, respectively. The above observations indicate that only transformer-based meteorological data and CNN-based remote sensing data analysis are underperforming.
\begin{figure}[htbp]
     \centering
     \begin{subfigure}[b]{0.31\linewidth}
         \centering
         \includegraphics[width=\linewidth]{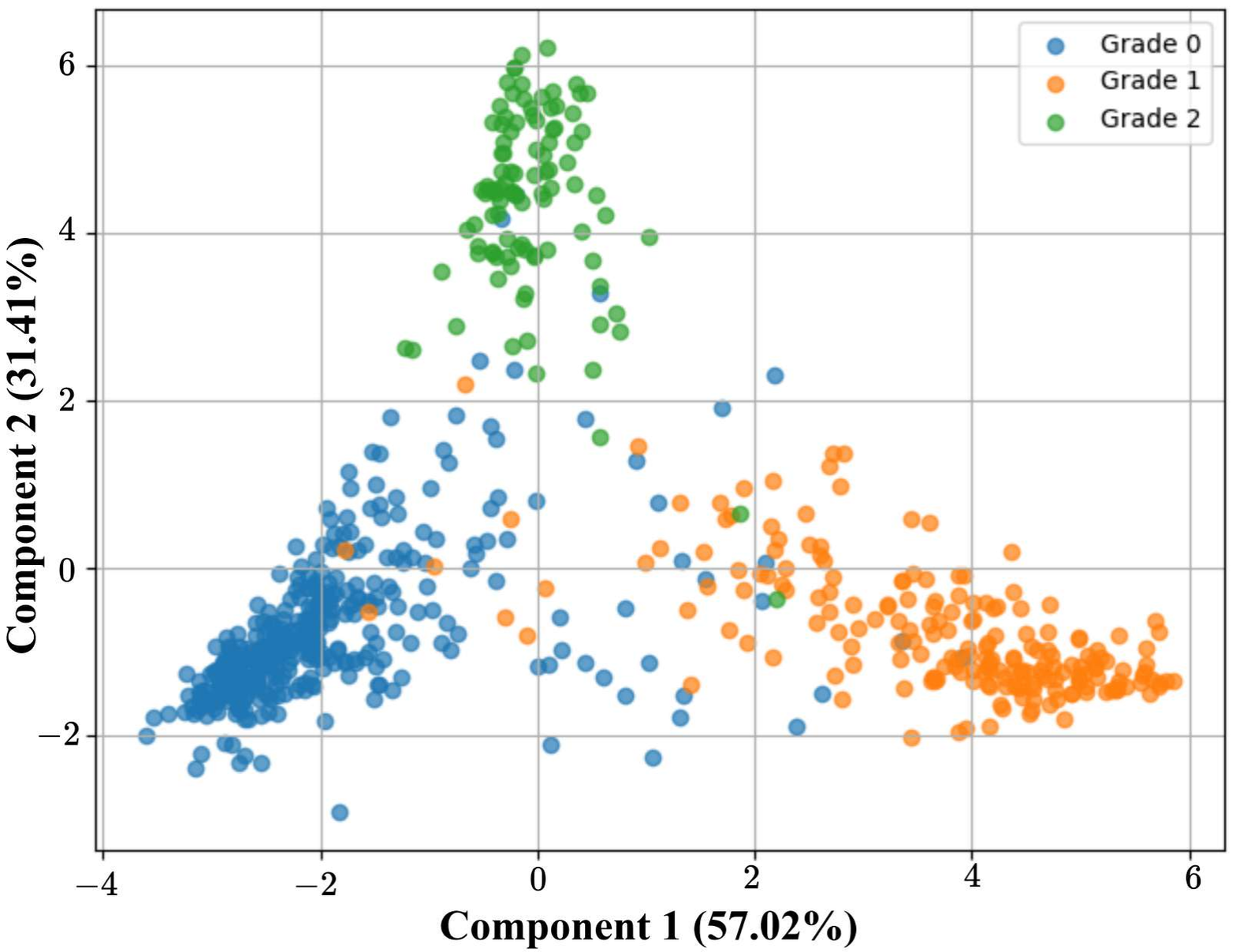}
         \caption{Distribution of features of the month of August}
         \label{13a}
     \end{subfigure}
     \hfill
     \begin{subfigure}[b]{0.31\linewidth}
         \centering
         \includegraphics[width=\linewidth]{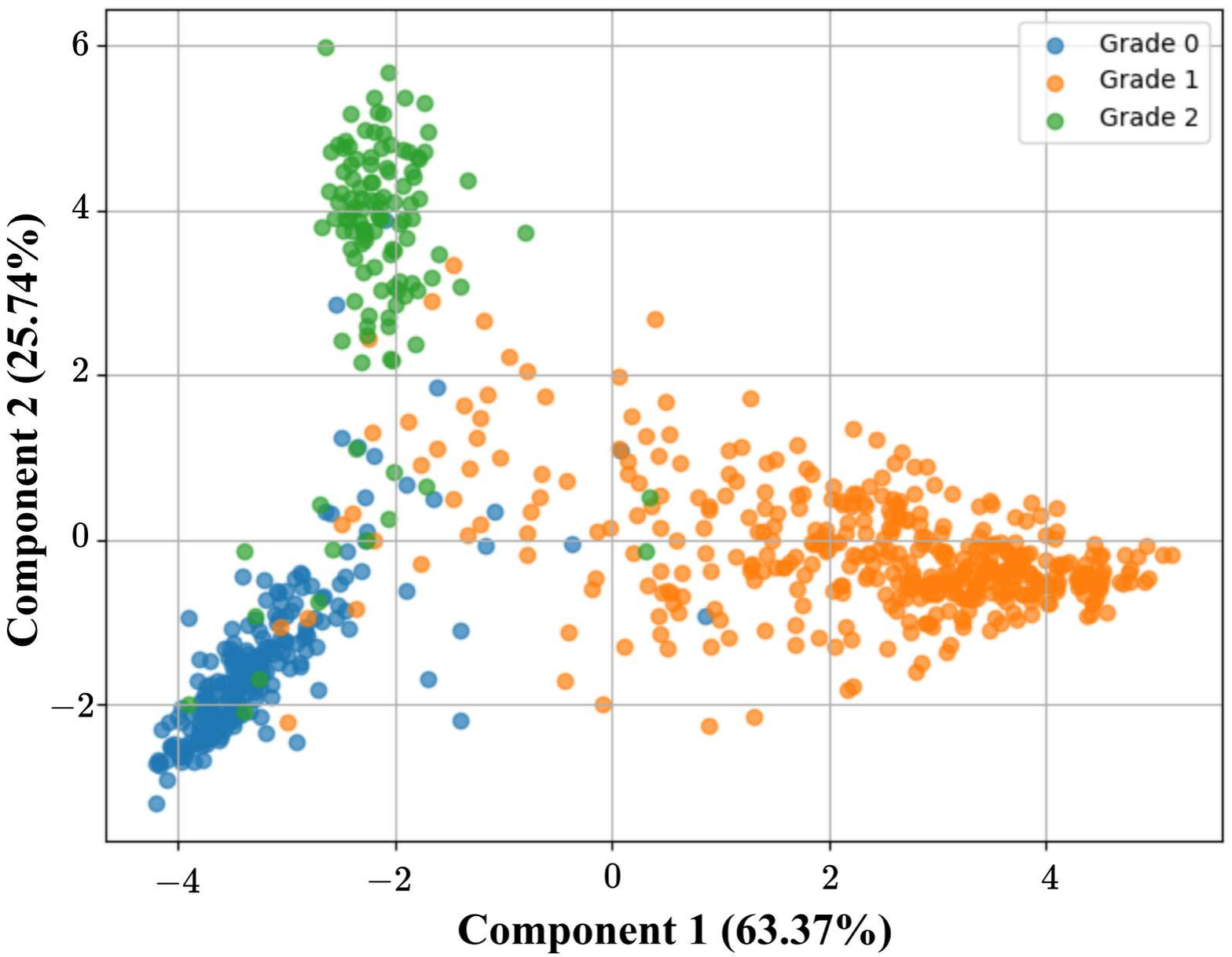}
         \caption{Distribution of features of the month of September}
         \label{13b}
     \end{subfigure}
     \hfill
     \begin{subfigure}[b]{0.31\linewidth}
         \centering
         \includegraphics[width=\linewidth]{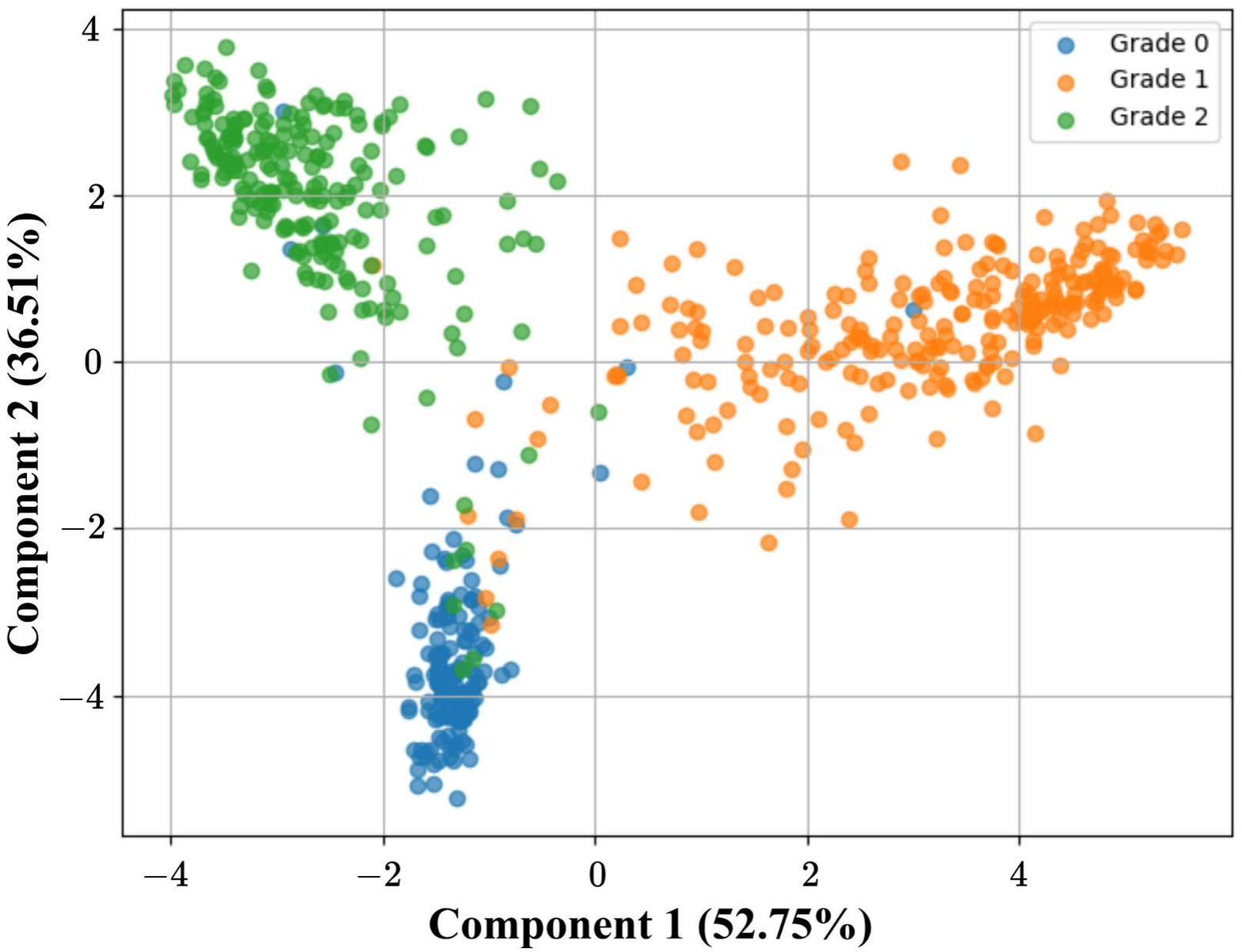}
         \caption{Distribution of features of  the month of October}
         \label{13c}
     \end{subfigure}
        \caption{Distribution of features of the proposed model for different months.}
        \label{fig 13}
\end{figure}
\subsection{Hyperparameter Sensitive Analysis}
\begin{figure}[!h]
    \centering
    \includegraphics[width=0.8\textwidth]{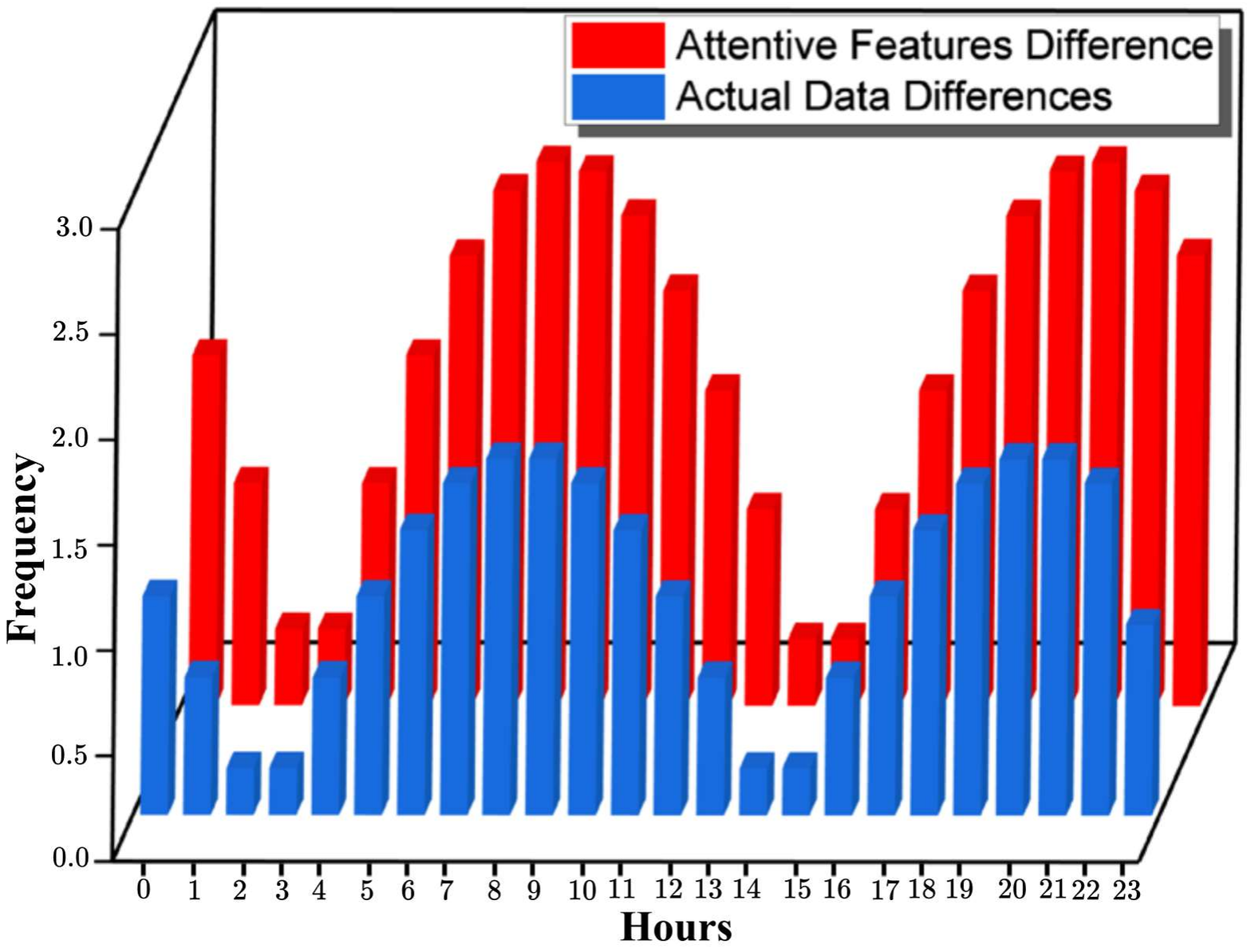}
    \caption{Observations between two consecutive hours of actual data and attentive data.}
    \label{fig 14}
\end{figure}

\begin{figure}[!h]
    \centering
    \includegraphics[width=0.8\textwidth]{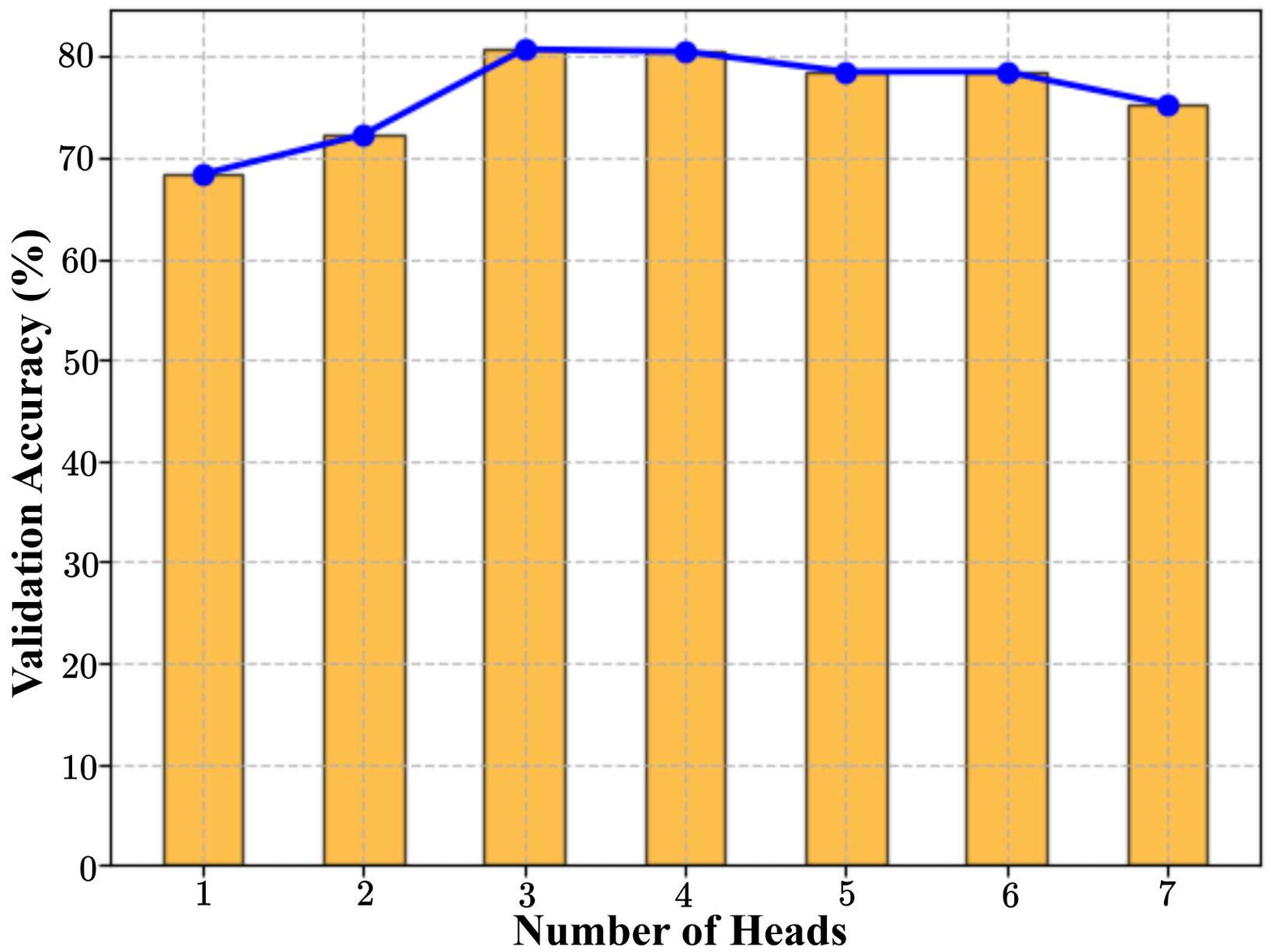}
    \caption{Performance of the attention mechanism with different numbers of heads.}
    \label{fig 15}
\end{figure}

The fusion model can handle complex data for disease assessment. To validate the model's effectiveness, its main components need to be examined. The fusion model is constructed from two main components: an attentive, positionally encoded transformer and a CNN-based GRU for modelling heterogeneous data. The transformer model consists of two main modules: an attention mechanism-enabled positional encoding and a transformer encoder. The attention mechanism is significant in assigning importance to each token (hourly data) before it is fed into the transformer encoder. Figure \ref{fig 14} shows that distinct differences are observed for consecutive hours with an attention mechanism.
The frequency of differences in the actual data is relatively low compared with that in attentive data. Hence, it is important to leverage the attention mechanism in the transformer network's positional encoding. Attention heads should be analyzed to finalize the attention mechanism. Different combinations of heads are investigated to develop the mechanism. Figure \ref{fig 15} depicts a frequency polygon-based bar plot to recognize the number of heads.\\ 
The results show that the approach's accuracy peaks at heads $3$ and $4$. Since increasing the number of heads is not feasible due to the model's complexity. Hence, the number of heads of the mechanism is selected as $3$. 
However, evaluation of the proposed model's modules in the framework, including quantities and entities, is required for observations.\\
\begin{table}[!htbp]
\caption{Experiments of the CNN-based GRU model on the Spectral dataset.}
\resizebox{\textwidth}{!}{
\begin{tabular}{|c|cccl|ccc|}
\hline
\multirow{2}{*}{\textbf{Model}} & \multicolumn{4}{c|}{\textbf{Components}}                                                                                                                                  & \multicolumn{3}{c|}{\textbf{Metrics}}                                                       \\ \cline{2-8} 
                       & \multicolumn{1}{c|}{\makecell{Convolution \\Layers}} & \multicolumn{1}{c|}{\makecell{Residual  \\ Networks}} & \multicolumn{1}{c|}{\makecell{Batch  \\ Normalization}} & \multicolumn{1}{c|}{\makecell{GRU \\Units}} & \multicolumn{1}{c|}{\makecell{Validation \\Accuracy\\(\%)}} & \multicolumn{1}{c|}{\makecell{Kappa\\($\kappa$)(\%)}} & \makecell{F$1$ Score\\ (\%)} \\ \hline
                    $A$   & \multicolumn{1}{c|}{$20$}                   & \multicolumn{1}{c|}{$5$}                    & \multicolumn{1}{c|}{$15$}                      &                    $1$            & \multicolumn{1}{c|}{$65.00$}                    & \multicolumn{1}{c|}{$66.13$}      & $66.40$           \\ \hline
                      $B$ & \multicolumn{1}{c|}{$30$}                   & \multicolumn{1}{c|}{$5$}                    & \multicolumn{1}{c|}{$15$}                      &   $1$                             & \multicolumn{1}{c|}{$68.60$}                    & \multicolumn{1}{c|}{$68.40$}      &  $69.60$          \\ \hline
                      $C$ & \multicolumn{1}{c|}{$40$}                   & \multicolumn{1}{c|}{$5$}                    & \multicolumn{1}{c|}{$15$}                      &   $1$                             & \multicolumn{1}{c|}{$73.35$}                    & \multicolumn{1}{c|}{$74.20$}      & $75.12$            \\ \hline
                      $D$ & \multicolumn{1}{c|}{$45$}                   & \multicolumn{1}{c|}{$5$}                    & \multicolumn{1}{c|}{$15$}                      &  $1$                              & \multicolumn{1}{c|}{$75.55$}                    & \multicolumn{1}{c|}{$76.20$}      &  $77.86$          \\ \hline
                      $E$ & \multicolumn{1}{c|}{$50$}                   & \multicolumn{1}{c|}{$5$}                    & \multicolumn{1}{c|}{$15$}                      &  $1$                              & \multicolumn{1}{c|}{$75.36$}                    & \multicolumn{1}{c|}{$76.20$}      &  $77.86$          \\ \hline
                      $F$ & \multicolumn{1}{c|}{$45$}                   & \multicolumn{1}{c|}{$10$}                    & \multicolumn{1}{c|}{$20$}                      &  $1$                              & \multicolumn{1}{c|}{$78.30$}                    & \multicolumn{1}{c|}{$78.80$}      &  $79.12$          \\ \hline
                      $G$ & \multicolumn{1}{c|}{$45$}                   & \multicolumn{1}{c|}{$15$}                    & \multicolumn{1}{c|}{$20$}                      &  $1$                              & \multicolumn{1}{c|}{$80.25$}                    & \multicolumn{1}{c|}{$81.35$}      &  $82.60$          \\ \hline
                      $H$ & \multicolumn{1}{c|}{$45$}                   & \multicolumn{1}{c|}{$20$}                    & \multicolumn{1}{c|}{$20$}                      &  $1$                              & \multicolumn{1}{c|}{$80.80$}                    & \multicolumn{1}{c|}{$82.80$}      &  $83.35$          \\ \hline
                      $I$ & \multicolumn{1}{c|}{$45$}                   & \multicolumn{1}{c|}{$15$}                    & \multicolumn{1}{c|}{$25$}                      &  $1$                              & \multicolumn{1}{c|}{$82.50$}                    & \multicolumn{1}{c|}{$83.38$}      &  $83.80$          \\ \hline
                      $J$ & \multicolumn{1}{c|}{$45$}                   & \multicolumn{1}{c|}{$15$}                    & \multicolumn{1}{c|}{$30$}                      &  $1$                              & \multicolumn{1}{c|}{$81.50$}                    & \multicolumn{1}{c|}{$82.80$}      &  $83.35$          \\ \hline
                      \textbf{K} & \multicolumn{1}{c|}{\textbf{45}}                   & \multicolumn{1}{c|}{\textbf{15}}                    & \multicolumn{1}{c|}{\textbf{25}}                      &  \textbf{2}                              & \multicolumn{1}{c|}{\textbf{84.45}}                    & \multicolumn{1}{c|}{\textbf{85.63}}      &  \textbf{86.32}          \\ \hline
                      $L$ & \multicolumn{1}{c|}{$45$}                   & \multicolumn{1}{c|}{$15$}                    & \multicolumn{1}{c|}{$25$}                      &  $3$                              & \multicolumn{1}{c|}{$84.45$}                    & \multicolumn{1}{c|}{$85.20$}      &  $85.86$          \\ \hline
\end{tabular}}
\label{tab 7}
\end{table}
Another module of the fusion approach is a CNN-based GRU network. The main components of the module are the proposed CNN network and GRU layers. To validate the effectiveness of the proposed CNN, its components are evaluated on the spectral dataset. The components of the CNN are convolution layers, residual networks, and batch normalization. Various network configurations need to be examined before implementing the model. Table \ref{tab 7} demonstrates the overall performance of the CNN-based GRU network in the spectral dataset.\\
The best performance is obtained for the model combination $K$, which consists of $45$ convolution layers, $15$ residual networks, $25$ batch normalizations, and $2$ GRU units. It shows that the combination offers the best approach, with the highest validation accuracy, Kappa ($\kappa$), and F$1$ score. Other combinations also achieved the same validation accuracy, but they missed the opportunity to match the performance in Kappa ($\kappa$) and F$1$ scores. The proposed version (Model $K$ in Table \ref{tab 7}) surpasses all other variants in all three performance criteria. The analysis in this study helps organize all modules by their importance and quantity to develop a fusion model.  
\subsection{Error Analysis}
\begin{figure}[!h]
    \centering
    \includegraphics[width=0.65\textwidth]{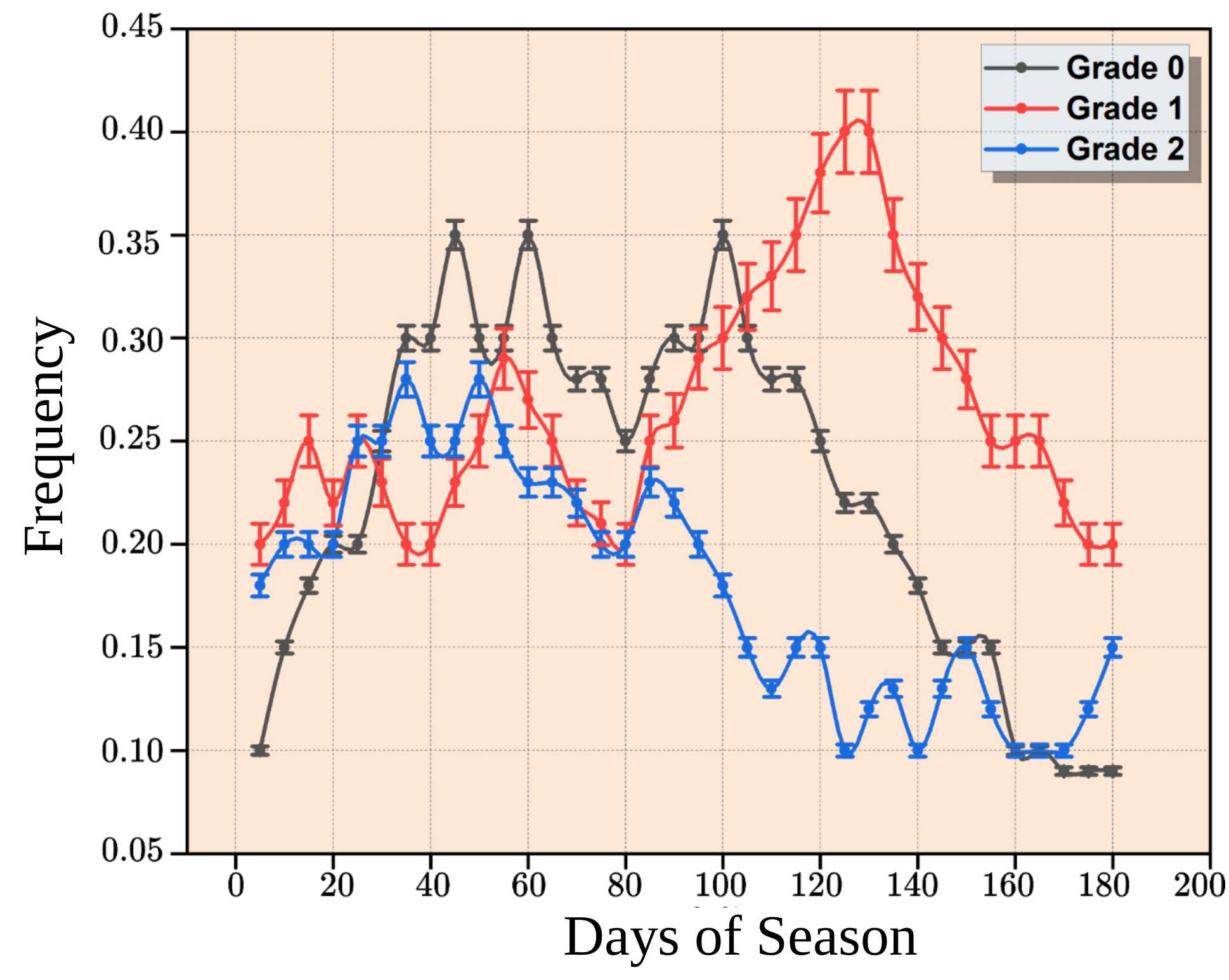}
    \caption{Mean error findings for each grade of $5$ day intervals of the season at different places.}
    \label{fig 16}
\end{figure}
This section analyses the improvement of the fused technique relative to ground observations, evaluates the resilience of the proposed fusion model across various scales, and assesses the influence of the original results on the predictions. A mean error analysis of the 2023 rice-growing season was conducted for specific locations. It is observed that the error has decreased significantly. However, the error of grade $1$ is increased from the $100^{th}$ day of the season to the $130^{th}$ day of the season. Maximum mean error frequency occurred $0.40$ for grade $1$, $0.35$ for grade $0$ and $0.28$ for grade $2$. The observation is visualized in Figure \ref{fig 16}.
In the context of mathematical statistics, residuals are the discrepancies between estimated and actual values, calculated from the classification probabilities in this study. The residuals are estimated using RMSE (root mean square error) and $R^2$ values. The performance of the model is shown in three time frames for every month of the rice growing season (Table \ref{tab 8}).\\
Table \ref{tab 8} shows $R^2$ values ranging from $0.85$ to $0.93$ for different timeframes. The model's RMSE lies between $0.16$ and $0.24$ for the same time. The maximum RMSE is $0.25$, and the lowest $R^2$ is $0.85$ for mid-September of $2023$. The model might be affected due to fluctuations in meteorological and spectral data. The average $R^2$ and RMSE were $0.89$ and $0.18$, respectively. The results show that the model is stable and credible in estimating early rice blast disease.
\begin{table*}[!ht]
\caption{Average $R^2$ and RMSE of the proposed method in all different areas ($2023$).}
\resizebox{\textwidth}{!}{
\begin{tabular}{|c|c|c|c|c|c|c|c|c|c|}
\hline
 \textbf{Metrics} & \textbf{\makecell{Early \\August}} & \textbf{\makecell{Mid \\August}} & \textbf{\makecell{Late\\ August}} & \textbf{\makecell{Early \\September}}  & \textbf{\makecell{Mid \\September}} & \textbf{\makecell{Late \\ September}} & \textbf{\makecell{Early\\October}} & \textbf{\makecell{Mid \\ October}} & \textbf{\makecell{Late\\October}} \\ \hline
 $R^2$& $0.92$ & $0.90$ & $0.87$ & $0.89$ & $0.85$ & $0.88$ & $0.89$ & $0.91$ & $0.93$ \\ \hline
 RMSE& $0.18$ & $0.24$ & $0.23$ & $0.21$ & $0.25$ & $0.23$ & $0.20$ & $0.18$ & $0.16$ \\ \hline
\end{tabular}}
\label{tab 8}
\end{table*}
\subsection{Ablation Study}
An ablation study is also conducted to understand the significance of the proposed modules in the model. Table \ref{tab 9} describes the utility of each module to boost the performance of the model. The attentive positional encoder mechanism improves the model's validation accuracy by $4.22\%$. However, other modules, including CNN and GRU, are also evaluated to assess the model's performance. It is observed that models based solely on transformers do not achieve satisfactory results. Therefore, CNN and GRU modules are also implemented and achieve higher performance of $3.73\%$, $2.13\%$, and $1.72\%$ for Validation accuracy, Kappa ($\kappa$), and F$1$ score, respectively. However, the study reveals that all module-based approaches pose significant repercussions for early rice blast prediction. It is observed that the proposed framework achieves $96.93\%$, $95.20\%$, and $95.79\%$, of validation accuracy, Kappa ($\kappa$) and  F$1$ scores, respectively.
\begin{table}[!htbp]
\caption{Performance analysis of different configurations of the proposed model.}
\label{tab 9}
\centering
\begin{tabular}{|c|ccc|ccc|}
\hline
\multirow{2}{*}{\textbf{Model}} & \multicolumn{3}{c|}{\textbf{Components}}                                                                                                          & \multicolumn{3}{c|}{\textbf{Metrics}}                                                                            \\ \cline{2-7} 
                       & \multicolumn{1}{c|}{\makecell{Attention in\\ Positional\\ Encoding}} & \multicolumn{1}{c|}{\makecell{CNN}} & \makecell{GRU} & \multicolumn{1}{c|}{\makecell{Validation \\Accuracy\\(\%)}} & \multicolumn{1}{c|}{\makecell{Kappa\\ ($\kappa$)(\%)}} & \multicolumn{1}{l|}{\makecell{F$1$ Score\\ (\%)}} \\ \hline
                      $A$ & \multicolumn{1}{c|}{\ding{55}}                                 & \multicolumn{1}{c|}{\ding{55}}                                &                     \ding{55}      & \multicolumn{1}{c|}{$76.50$}                    & \multicolumn{1}{c|}{$77.83$}      & $77.50$                                \\ \hline
                      $B$ & \multicolumn{1}{c|}{\ding{51}}                                 & \multicolumn{1}{c|}{\ding{55}}                                &                 \ding{55}          & \multicolumn{1}{c|}{$80.72$}                    & \multicolumn{1}{c|}{$83.50$}      &  $84.60$                               \\ \hline
                      $C$ & \multicolumn{1}{c|}{\ding{55}}                                 & \multicolumn{1}{c|}{\ding{51}}                                &             \ding{55}              & \multicolumn{1}{c|}{$78.60$}                    & \multicolumn{1}{c|}{$79.20$}      &      $78.00$                           \\ \hline
                      $D$ & \multicolumn{1}{c|}{\ding{55}}                                 & \multicolumn{1}{c|}{\ding{51}}                                &          \ding{51}                 & \multicolumn{1}{c|}{$84.45$}                    & \multicolumn{1}{c|}{$85.63$}      &   $86.32$                              \\ \hline
                      $E$ & \multicolumn{1}{c|}{\ding{51}}                                 & \multicolumn{1}{c|}{\ding{51}}                                &          \ding{51}                 & \multicolumn{1}{c|}{\textbf{96.93}}                    & \multicolumn{1}{c|}{\textbf{95.20}}      &   \textbf{95.79}                              \\ \hline
\end{tabular}
\end{table}
\subsection{Model Complexity}
To evaluate the effectiveness and computational efficiency of the models developed to differentiate among several disease grades, the number of parameters, FLOPs (Floating-Point Operations Per Second) and prediction time were examined. The model comprises $2.15$ million trainable parameters and $3.01$ GFLOPs, as shown in Table \ref{tab 4}. The proposed model takes less than $1$ second per sampling, with discrepancies ranging from $15$ ms to $24$ ms. The model accurately predicted the disease stages within an average of $20$ ms.

\subsection{Sustainable AI Assessment}
In evaluating the deployment viability of the proposed framework for early rice blast recognition, computational sustainability metrics—including operational electricity consumption, carbon dioxide equivalent ($CO_2eq$) emission rates, and hardware energy dissipation—are examined alongside predictive accuracy. High-performance deep learning models often scale their parameter count quadratically, posing a trade-off between macro-level classification precision and energy consumption. Designing a lightweight residual CNN paired with a highly constrained temporal transformer loop ($T=2$), our proposed architecture caps memory demands at approximately $2.15\text{~M}$ parameters and restricts computational complexity to $3.01\text{~GFLOPs}$ per inference cycle. To quantify the physical footprint, a benchmark was conducted on a standard edge-station hardware configuration with an active thermal design power (TDP) profile. The cumulative electricity consumption ($E$) and corresponding carbon footprint are formalized via the operational grid carbon intensity factor:

\begin{equation}
E = \frac{T \times \left( \text{TDP}_{\text{GPU}} \times \mu_{\text{GPU}} + \text{TDP}_{\text{CPU}} \times \mu_{\text{CPU}} \right) \times \text{PUE}}{1000}
\end{equation}

where $T$ represents processing duration, $\mu$ represents mean utilization factors, and $\text{PUE}$ is the power usage effectiveness coefficient clamped at a local baseline of $1.0$.
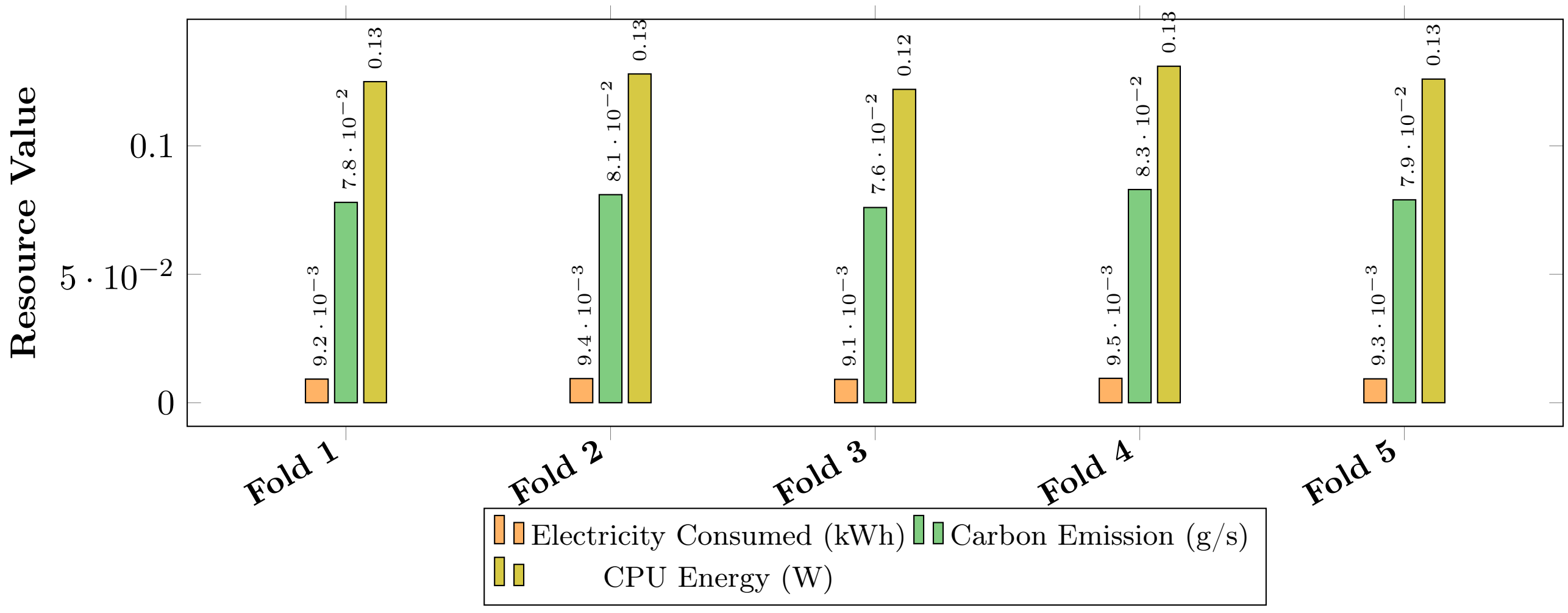
\begin{figure}[!htbp]
\centering
\begin{tikzpicture}
\begin{axis}[
    ybar,
    enlargelimits=0.15,
    legend style={at={(0.5,-0.2)}, anchor=north, legend columns=2, font=\footnotesize},
    ylabel={\textbf{Resource Value}},
    symbolic x coords={Fold 1,Fold 2, Fold 3,Fold 4, Fold 5},
    xtick=data,
    nodes near coords,
    nodes near coords align={vertical},
    every node near coord/.append style={font=\tiny, rotate=90, anchor=west},
    x tick label style={rotate=30, anchor=east, font=\small\bfseries},
    width=\columnwidth,
    height=6cm,
    bar width=7pt
]
\addplot[fill=orange!60!white] coordinates {(Fold 1,0.0092) (Fold 2,0.0094) (Fold 3,0.0091) (Fold 4,0.0095) (Fold 5,0.0093)};
\addplot[fill=green!60!black!50!white] coordinates {(Fold 1,0.078) (Fold 2,0.081) (Fold 3,0.076) (Fold 4,0.083) (Fold 5,0.079)};
\addplot[fill=yellow!80!black] coordinates {(Fold 1,0.125) (Fold 2,0.128) (Fold 3,0.122) (Fold 4,0.131) (Fold 5,0.126)};

\legend{Electricity Consumed (kWh), Carbon Emission (g/s), CPU Energy (W)}
\end{axis}
\end{tikzpicture}
\caption{Sustainability assessment metrics of the Proposed Model across different cross-validation experimental trials, detailing resource consumption profiles.}
\label{fig:sustainability_plot}
\end{figure}

The empirical evaluation produces an average $CO_2eq$ emission rate of approximately $7.94 \times 10^{-3}\text{~g/s}$ for active image processing windows, representing a major footprint reduction compared to other frameworks that exceed standard grid thresholds. Given this efficiency window, the hybrid network effectively captures vital cues without exceeding local hardware constraints, validating its use as both a highly accurate and practically sustainable solution for real-world eco-friendly agricultural monitoring systems.
\subsection{Limitations and future scope}
This study has several limitations for early disease state recognition. The model may produce some erroneous results due to other environmental parameters. In case of erroneous results, the complex relationship of meteorological and spectral data affected the performance of the fusion model. This study focuses on limited regions, seasons, and parameters for disease analysis. However, input variations may affect the model's capability. The recorded hourly temperature could be more useful for the sequential analysis. We are focused on recording and generating data using algorithms to improve user effectiveness, which need to be restructured utilizing learning algorithms. Spectral data can lead to misinterpretation when analyzed in the presence of unavailable or noisy pixels. Further studies will address missing and noisy values to improve spectral representations, enhancing the model's performance. 

Further research is needed to evaluate the model's effectiveness in different nations and locations with multi-modal satellite observations and parameters.
\section{Conclusion}\label{sec 5}
This paper presents a resource-efficient multimodal fusion framework for early rice blast forecasting at the field scale, supporting sustainable agriculture through timely disease intervention and optimized pest management. The proposed framework integrates an attentive positional encoding-based Transformer with a lightweight CNN-GRU architecture to jointly model heterogeneous meteorological and remote sensing data. The attention-enhanced positional encoding captures long-term temporal dependencies in meteorological observations, while the CNN-GRU learns complementary spectral–temporal representations from multi-temporal remote sensing imagery. Their fusion effectively models disease progression and environmental interactions, enabling reliable forecasting $5$\textendash$10$ days before symptom onset.

Extensive experiments across multiple agro-climatic zones demonstrate the effectiveness of the proposed framework, achieving a validation accuracy of $96.93\%$ and an $F1$-score of $95.79\%$. Ablation studies validate the contribution of each module, while complexity analysis confirms the computational efficiency of the architecture. The framework requires only $2.15$ million trainable parameters and $3.01$ $GFLOP$s, achieving an average inference time of approximately $20$ $ms$ with a carbon emission rate of only $7.94 \times 10^{-3}gCO_2 eq/s$. These results demonstrate that high predictive accuracy can be achieved alongside low computational and environmental costs, providing a practical foundation for resource-efficient AI, sustainable disease surveillance, and intelligent decision support in precision agriculture. Future work will focus on improving model generalization across diverse agro-climatic regions, incorporating additional sensing modalities, and further enhancing the energy efficiency of multimodal learning for sustainable agricultural systems.

\bibliographystyle{IEEEtran}
\bibliography{ref.bib}
\end{document}